\documentclass[runningheads]{llncs}
\usepackage{graphicx}
\usepackage{amsmath,amssymb} 
\usepackage{color}
\usepackage{booktabs}
\usepackage{mathtools}
\usepackage{subcaption}
\graphicspath{ {images/} }

\newcommand\blfootnote[1]{%
  \begingroup
  \renewcommand\thefootnote{}\footnote{#1}%
  \addtocounter{footnote}{-1}%
  \endgroup
}

\begin{document}

\title{Face Completion with Semantic Knowledge \\ and Collaborative Adversarial Learning} 
\titlerunning{Face Completion with Collaborative Adversarial Learning} 


\author{Haofu Liao\inst{1}\textsuperscript{*} \and
Gareth Funka-Lea\inst{2} \and
Yefeng Zheng\inst{3}\textsuperscript{*} \and
Jiebo Luo\inst{1} \and \\
S. Kevin Zhou\inst{4}\textsuperscript{*}}
%

\authorrunning{Liao et al.} 


\institute{Department of Computer Science, University of Rochester, Rochester, USA \email{hliao6@cs.rochester.edu}\and
Digital Technology and Innovation, Siemens Healthineers, Princeton, USA
\and Tencent X-Lab, Shenzhen, China
\and Institute of Computing Technology, Chinese Academy of Sciences, Beijing, China\\
}

\maketitle

\begin{abstract}\blfootnote{\textsuperscript{*}The work was done when the authors were with Siemens Healthineers.}Unlike a conventional background inpainting approach that infers a missing area from image patches similar to the background, face completion requires \textit{semantic knowledge} about the target object for realistic outputs. Current image inpainting approaches utilize generative adversarial networks (GANs) to achieve such semantic understanding. However, in adversarial learning, the semantic knowledge is learned implicitly and hence good semantic understanding is not always guaranteed. In this work, we propose a \textit{collaborative adversarial learning} approach to face completion to explicitly induce the training process. Our method is formulated under a novel generative framework called collaborative GAN (collaGAN), which allows better semantic understanding of a target object through collaborative learning of multiple tasks including face completion, landmark detection and semantic segmentation. Together with the collaGAN, we also introduce an inpainting concentrated scheme such that the model emphasizes more on inpainting instead of autoencoding. Extensive experiments show that the proposed designs are indeed effective and collaborative adversarial learning provides better feature representations of the faces. In comparison with other generative image inpainting models and single task learning methods, our solution produces superior performances on all tasks.

\end{abstract}

\section{Introduction}

Image inpainting is the process of reconstructing a missing region in an image such that the inpainted area is visually consistent with its neighboring pixels and the inpainted image overall looks realistic. Traditional approaches to this problem either require that the filling information is available in the image \cite{he2012statistics,huang2014image,barnes2009patchmatch} or rely on the availability of a large photo database to retrieve the missing region \cite{whyte2009get,hays2007scene}. These approaches work well for the \textit{background inpainting} problem, i.e., filling the missing part with image patches similar to its background, as inpainting can be performed through pattern matching. However, when it comes to complete a missing part of an object where no existing patches can be matched or retrieved, the traditional approaches may fail. For example, if a mouth is missing, it is not possible to synthesize the mouth using image patches from other face parts. Instead, image inpainting in this case requires \textit{semantic knowledge} about faces, e.g., location, shape, color and texture of face parts.
    
\begin{figure}[t]
  \centering
  \begin{subfigure}[b]{0.15\textwidth}
      \includegraphics[width=1.03\linewidth]{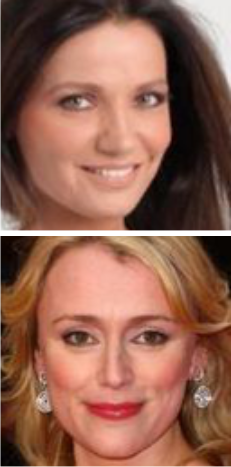}
      \vspace*{-14pt}
      \caption{}
  
  \end{subfigure}
  \begin{subfigure}[b]{0.15\textwidth}
      \includegraphics[width=1.03\linewidth]{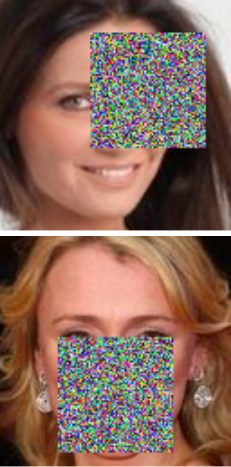}
      \vspace*{-14pt}
      \caption{}

  \end{subfigure}
  \begin{subfigure}[b]{0.15\textwidth}
      \includegraphics[width=1.03\linewidth]{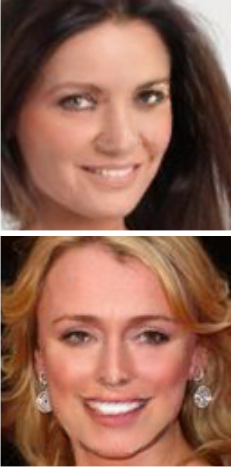}
      \vspace*{-14pt}
      \caption{}

  \end{subfigure}
  \begin{subfigure}[b]{0.15\textwidth}
      \includegraphics[width=1.03\linewidth]{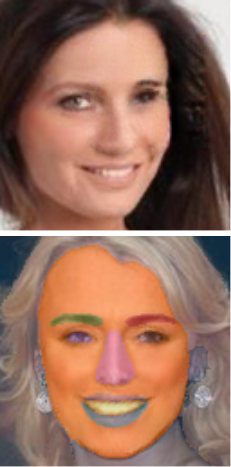}
      \vspace*{-14pt}
      \caption{}
  \end{subfigure}
  \caption{Inpainting results using our method. (a) original image. (b) masked image. (c) inpainted image with collaborative learning. (d) inpainted image without collaborative learning (top) and segmentation mask from our method superimposed on the inpainted image (bottom).}
  \label{fig: collaborative_inpainting}
\end{figure}

To address this \textit{object completion} problem in image inpainting, recent models \cite{pathak2016context,yeh2016semantic,li2017generative} propose to use generative adversarial networks (GANs) for more semantically consistent results. However, for generative models, the semantic understanding is implicitly learned through adversarial training. There are no direct constraints on the structure of the target object and hence the inherent semantic understanding is not always guaranteed. Fortunately, in recent years, the success of deep learning has made the semantic labels of objects accessible. In this work, we investigate the possibility of introducing the semantic knowledge of face labels to the adversarial training of face completion for better induction of semantic understanding.


We focus on helping the inpainting model better understand the underlying structure of faces through the collaborative learning of other face related tasks. We argue that current approaches using generative inpainting models alone may not be able to produce structurally realistic results in some cases. For example, when an eye is missing from the image, the inpainting model should be able to predict the missing eye's location and shape based on the facial symmetry. However, as shown in the first row of Figure \ref{fig: collaborative_inpainting}, the generative image inpainting model trained without using our proposed collaborative method produces a structurally unrealistic face (Figure \ref{fig: collaborative_inpainting}(d) top) with the inpainted eye smaller and darker than the eye outside the corrupted region. In contrast, a collaboratively trained model can keep the structural consistency between the inpainted region and the nearby context (Figure \ref{fig: collaborative_inpainting}(c) top). In addition, we also find that models trained in this manner tend to produce visually consistent results among tasks. As demonstrated in Figure \ref{fig: collaborative_inpainting}(d) bottom, the segmentation result is closely aligned with the inpainting result other than the ground truth. This provides  a clear evidence that they are inherently helping each other during training and the knowledge is shared instead of individually learned.

To this end, we propose an innovative image-to-image generative network for face completion. The proposed method formulates a collaborative GAN to facilitate the direct learning of multiple tasks. For the generator, the network outputs multiple channels for each task and has them share most of the network parameters for better collaborative learning. We also stand apart from the existing inpainting models by introducing skip connections between the encoder and decoder \cite{ronneberger2015u,isola2016image}. For the discriminators, we apply conditional GAN (cGAN) \cite{mirza2014conditional} for better transformation quality and have dedicated discriminators for each task. For the loss function, we introduce an inpainting concentrated scheme to allow the model focusing on the inpainting itself instead of autoencoding the context. Our experimental results demonstrate the effectiveness of the proposed design and better feature representations can be obtained with the proposed collaborative GAN. Comparing with other generative models without using collaborative adversarial learning, our approach consistently produces remarkably more realistic inpainting results. Comparing with single task adversarial learning, our joint approach produces better performances on all tasks.

\section{Related Work}
    
\noindent \textbf{Generative Image to Image Transformation} \hspace{5pt} Image inpainting can be seen as a special case of the image-to-image transformation problem in that image inpainting tries to transform a cropped image to a reconstructed one.
One of the typical image-to-image transformation problems is autoencoding \cite{hinton2006reducing}.
In relation to inpainting, a seminal work in this area is the denoising autoencoder \cite{vincent2008extracting}. It introduces an inpainting-like scheme to autoencoding, hoping the autoencoder can learn a better feature representation by recovering the damage to the input image. Another related work is \cite{larsen2015autoencoding}. It incorporates GANs into a variational autoencoder (VAE) \cite{kingma2013auto} and argues that the network trained using the adversarial loss and VAE loss can give a more representative feature vector of the input image.

For other image-to-image transformation problems, \cite{ledig2016photo} proposes to use GANs for image super-resolution. It has the standard generative image-to-image model setting: a generator that maps the low resolution input to high resolution and a discriminator that tells if the input image is a high resolution one or not. It also includes perceptual loss \cite{ledig2016photo} to further regularize the realism.
\cite{isola2016image} proposes a general framework for image-to-image translation. It improves the generative image-to-image networks by introducing image-conditional GANs and PatchGAN. Inspired by \cite{isola2016image}, \cite{li2018carigan} proposes a generative image-to-image framework for caricature image generation from face image. \cite{liao2018adversarial} applies generative image-to-image network for medical image artifact reduction.

\vspace{2mm}

\noindent \textbf{Semantic Inpainting} \hspace{5pt} The term \textit{semantic inpainting} is first introduced by context encoder (CE) \cite{pathak2016context} to address the challenging inpainting case where a large region of the image is missing and the inpainting generally requires semantic understanding of the context. The paper proposes to learn a better autoencoder by having it recover the missing part of an input image. To our best knowledge, it is the first work that introduces adversarial loss into the inpainting problem. Another approach to semantic inpainting is \cite{yeh2016semantic}. The method is based on a pretrained GAN model that maps a noise vector to the generator manifold. The algorithm finds the noise vector such that the generated image through the pretrained GAN model minimizes both a contextual loss and a perceptual loss. This work is an indirect approach to image inpainting. Due to the limitation of the pretrained GAN model, it may fail to produce good results when the image resolution is high or the scene is complex. \cite{yang2017high} proposes a multi-scale neural patch synthesis network for high-resolution image inpainting. Since our work focus on the benefit of collaborative adversarial training, high-resolution image inpainting is not within the cope of this study. In fact, our work can be easily extended to high-resolution settings for better performance.

The closest work to ours is \cite{li2017generative}. It advances the state of the art  \cite{pathak2016context} by introducing a parsing network to further regularize the inpainting through semantic parsing/segmentation and a local discriminator to ensure the generated contents are semantically coherent. However, the parsing network is pretrained and independent of the generator. Thus, the semantic parsing information is not shared with the generator during training. The semantic parsing loss is also not applied directly to the generator. As a result, the accuracy of computing such a loss is limited by the parsing network which further limits the final performance of the generator. Meanwhile, for the local discriminator, it requires the mask to be rectangular and hence cannot be generalized to other inpainting cases such as noise inpainting. \cite{iizuka2017globally} also proposes a similar global and local discriminator design and it suffers from the same disadvantage as \cite{li2017generative}. Semantic inpainting has also been applied to address medical problems. \cite{liao2018more} introduces a 3D inpainting task to improve the performance of ultrasound image segmentation. \cite{liao2019generative,lin2019dudonet} convert the metal artifact reduction problem to the X-ray image inpainting problem and the metal artifacts are removed by reconstructing from the inpainted X-ray images.

\vspace{2mm} \noindent \textbf{Multi-Task Learning} \hspace{5pt} Multi-task learning (MTL) \cite{caruana1998multitask} refers to the process of learning multiple tasks jointly in order to improve the generalization performance of the model. It has been widely used in deep neural networks (DNNs) for various tasks, such as object detection \cite{girshick2015fast}, action recognition \cite{NIPS2014_5353}, landmark detection \cite{zhang2014facial}, etc. In terms of generative models, many works have introduced adversarial learning in a multi-task fashion to improve the learning of the main tasks. \cite{ganin2016domain} uses adversarial learning for better domain adaption of the main task. \cite{li2017perceptual} proposes a perceptual GAN that learns super-resolution together with object detection for better performance of small object detection. \cite{liu2017adversarial} leverages GANs to generate shared features that are independent of different text classification tasks. In this work, we contribute to the literature with a novel image-to-image generative framework that collaboratively learns multiple tasks for better semantic understanding and, ultimately, yields better image inpainting.

\begin{figure*}[t]
    \centering
    \includegraphics[width=\textwidth]{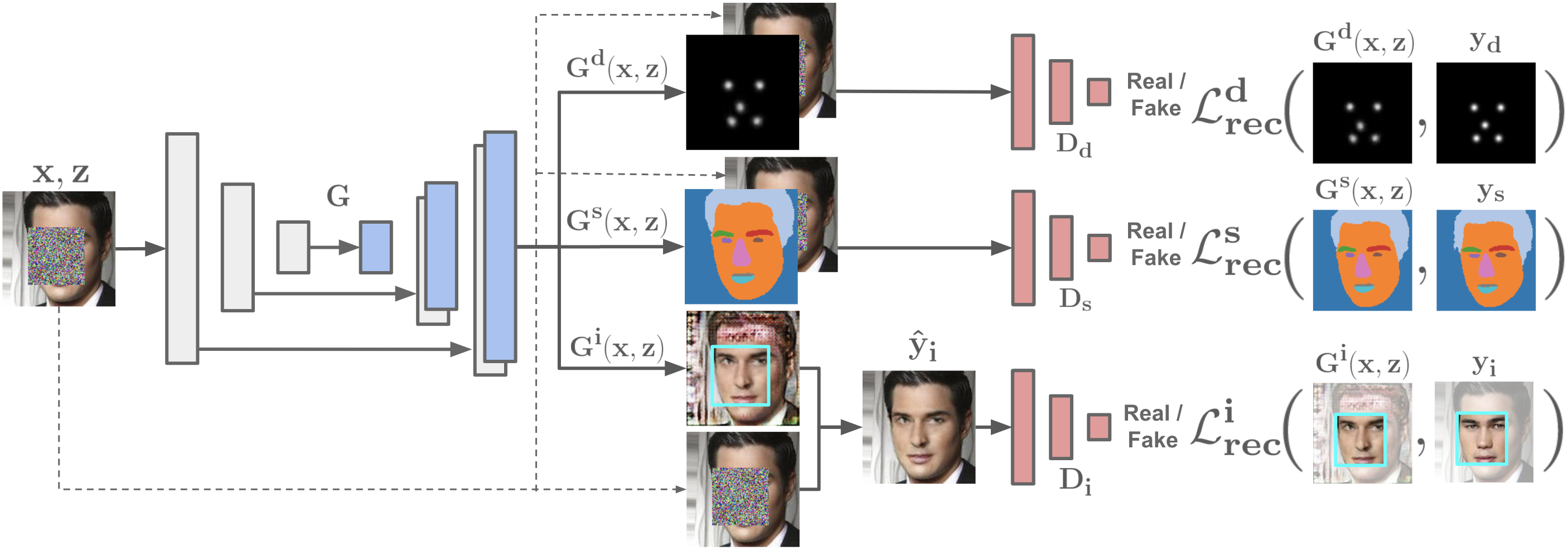}
    \caption{The architecture of the proposed method. The network is trained collaboratively with three tasks: inpainting $i$, segmentation $s$ and (landmark) detection $d$. The generator $G$ takes a masked image $x$ as input and outputs the inpainted image $G^i(x,z)$, segmentation mask $G^s(x,z)$ and detection heatmap $G^d(x,z)$ simultaneously. The discriminators $D_i$, $D_s$ and $D_d$ are used for adversarial learning. In addition, reconstruction losses $\mathcal{L}_{rec}^i$, $\mathcal{L}_{rec}^s$ and $\mathcal{L}_{rec}^d$ are also applied to the three tasks, respectively.}
    \label{fig: architecture}
\end{figure*}

\section{Collaborative Face Completion}

The proposed collaborative face completion method is formulated under a novel GAN framework which we call \textit{collaborative GAN} (collaGAN). The proposed framework aims to inductively improve the main generation task (face completion in our case) by incorporating the additional knowledge embedded in other tasks. In this section, we give the formal definition of collaGAN and its connection to face completion.

\subsection{Collaborative GAN}

Let $\Omega$ be a finite set of tasks and $y_t$ be a data sample of task $t \in \Omega$. A collaGAN learns a mapping from an input image $x$ and a random noise $z$ to a set of outputs $\{y_{t} | t \in \Omega \}$, i.e., $G : \{x, z\} \rightarrow\{y_{t} | t \in \Omega \}$. The network is trained in an adversarial fashion. The generator $G$ tries to generate data samples as ``real'' as possible such that, $ \forall t \in \Omega$, the adversarially trained discrimintor $D_t$  cannot tell if a sample is generated by $G$ or from the data domain of task $t$. 

Similar to the classic GAN, the objective function of a collaGAN can be given as follows:
\begin{equation} 
    \begin{split}
    &\mathcal{L}^{\Omega}_{adv} = \min_G \max_{D_{\Omega}} \sum_{t\in\Omega}\mathbb{E}_{x,y_{t} \sim p_{data} (x, y_{t})}[\log D_t(x, y_t )] \\
    & + \lambda_{adv}^t\mathbb{E}_{x \sim p_{data} (x), z \sim p_z(z)}[1 - \log D_t(x, G^t(x, z))],
    \end{split}
\end{equation}
Here, $D_{\Omega}=\{D_t | t \in \Omega\}$. $p_{data}$ and $p_z$ denote the data and noise distribution, respectively. $G(x, z)$ is the channel-wisely stacked generator outputs of the tasks with $G^t(x, z)$ denoting the ouput generated for task $t$. $\lambda_{adv}^t$ balances the importance of the generator loss for task $t$. $D_{\Omega}$ and $G$ play a minmax game where $D_{\Omega}$ tries to maximize the objective and $G$ tries to minimize it.

In this work, as shown in Figure \ref{fig: architecture}, $\Omega = \{i, s, d\}$, denoting three tasks of \textit{inpainting, segmentation, and (landmark) detection}. $x$ is the occluded image, $y_{i}$ is the original face, $y_{s}$ is the segmentation mask and $y_{d}$ is the detection heatmap (See Section \ref{sec: datasets}). $D_{\Omega}=\{D_{i}, D_{s}, D_{d}\}$ where $D_{i}$, $D_{s}$ and $D_{d}$ are the discriminators for the inpainting, segmentation and detection tasks, respectively. $G(x, z) = [G^i(x, z), G^s(x, z), G^d(x, z)]$ are the generator outputs for the three tasks.

Note that instead of only applying adversarial loss to the inpainted image, we have dedicated discriminators for each of the tasks. Our design follows the observation that GANs can also be helpful in some non-generative tasks \cite{luc2016semantic}. In our case, GANs are used to keep the long-range spatial label contiguity of the detection heatmap and the segmentation mask. We have also experimented with an optional setting that feeds the outputs of all the tasks into a single discriminator and discover that the single discriminator selectively ignores the inpainting result as the outputs from other tasks are much easier for judging the realness. Besides, the collaGAN is conditioned on multiple discriminators, one per task. Such a design, on one hand, ensures the perceptual quality of the generated image and, on the other hand, keeps the spatial consistency between the input image and generated image. This choice was also shown to be effective in \cite{isola2016image}.

\subsection{Reconstruction Loss}

It has been found by previous approaches \cite{isola2016image,pathak2016context} that mixing the adversarial loss with a reconstruction loss is beneficial to generative image-to-image models. A reconstruction loss gives pixel-level measurement of the errors which is a direct regularization between the output and the ground truth. It can capture the overall structure of the target object but is usually unable to give a sharp output. An adversarial loss, on the other hand, can produce perceptually better result but the output may not be structurally consistent with input. Therefore, we combine these two to achieve both realistic and coherent output.

In our case, the reconstruction loss is computed for the three tasks as denoted in Figure \ref{fig: architecture}. For the inpainting task, the reconstruction loss $\mathcal{L}_{rec}^i$ measures the L1 distance between the inpainted image and the unoccluded image. Here, we use the L1 loss instead of the L2 loss due to the observation that the L2 loss tends to give slightly blurry outputs. For the segmentation mask, we first convert the label mask to a multi-channel map with each channel denoting the binary mask of a label. Then, we apply the L2 loss $\mathcal{L}_{rec}^s$ to the multi-channel map to measure the difference between the network output and the ground truth. We use the L2 loss against the typical cross-entropy loss simply for the ease of implementation as the generator (See section \ref{sec: network_architecture}) produces outputs with values between $(-1, 1)$ which favors a regression loss. In our experiments, we find the L2 loss gives good enough segmentation map for this study. For the detection heatmap, we also use the L2 loss $\mathcal{L}_{rec}^d$ as the regularizer adapting the choices from \cite{pfister2015flowing,payer2016regressing}. Let $\Omega = \{i, s, d\}$, the total reconstruction loss can be written as
\begin{equation}
    \mathcal{L}^{\Omega}_{rec} = \sum_{t \in \Omega} \lambda_{rec}^t \mathcal{L}_{rec}^{t},
\end{equation}
where $\mathcal{L}_{rec}^{t}$ denotes the reconstruction loss of task $t$ and $\lambda_{rec}^t$ denotes the importance of each loss. The final objective function is then given by
\begin{equation}
    \mathcal{L}^{\Omega} = \mathcal{L}_{adv}^{\Omega} + \mathcal{L}_{res}^{\Omega}.
\end{equation}


\begin{figure}[t]
    \centering
    \begin{subfigure}[b]{0.15\textwidth}
        \includegraphics[width=1.03\linewidth]{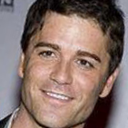}
        \vspace*{-14pt}
        \caption{}
    \end{subfigure}
    \begin{subfigure}[b]{0.15\textwidth}
        \includegraphics[width=1.03\linewidth]{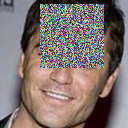}
        \vspace*{-14pt}
        \caption{}
    \end{subfigure}
    \begin{subfigure}[b]{0.15\textwidth}
        \includegraphics[width=1.03\linewidth]{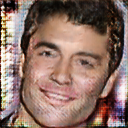}
        \vspace*{-14pt}
        \caption{}
    \end{subfigure}
    \begin{subfigure}[b]{0.15\textwidth}
        \includegraphics[width=1.03\linewidth]{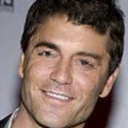}
        \vspace*{-14pt}
        \caption{}
    \end{subfigure}
    \caption{Example output of the network trained with the inpainting concentrated scheme. From (a) to (d): (a) original image, (b) masked image, (c) output from the generator, and (d) the final inpainting result by combining (b) and (c).}
    \label{fig: inpainting_concentration}
\end{figure}

\subsection{Inpainting Concentrated Generation} \label{sec: inpainting_concentrated_generation}


In previous work \cite{li2017generative}, image inpainting is performed through an autoencoder and the unoccluded region is reconstructed along with the occluded part. Thus, the network spends significant portion of its computing power on autoencoding the already available information while the inpainting itself is not fully addressed. A direct approach \cite{pathak2016context} to this problem is having the generator only outputs the content within the mask. However, this only works for the case where the masks are rectangular. When random shaped regions are occluded, this approach fails as convolutional neural networks cannot generate non-rectangular images. To address this problem for arbitrary shaped masks, we propose an inpainting concentrated scheme. The scheme consists of two parts: an adversarial part and a reconstruction part. For the adversarial part, we modify the adversarial loss for the inpainting such that the discriminator, instead of judging the realness of the generated image, concentrates on finding the incoherence between the inpainted region and the context. Formally, let $M$ be a binary mask where pixels in the occluded region are 0 and anywhere else are 1. The new adversarial loss for inpainting can be written as
\begin{equation}
    \begin{split}
    &\mathcal{L}_{adv}^{i} =  \mathbb{E}_{x,y_i \sim p_{data} (x, y_i)}[\log D_i(x, y_i)] \\
    & + \lambda_i\mathbb{E}_{x \sim p_{data} (x), z \sim p_z(z)}[ (1 - \log D_i(x, \hat{y}_i ))],
    \end{split}
\end{equation}
where
\begin{equation} \label{eq: replacing}
\hat{y}_i = G^i(x, z) \odot (1 - M) + x \odot M.
\end{equation}
As demonstrated in Figure \ref{fig: architecture}, $\hat{y}_i$ is nothing but the inpainting output from the generator with the unoccluded region replaced by the ground truth. Such a replacement guarantees that the discriminator does not need to worry about the unrealness of the context and thus the inpainted region is concentrated. For the reconstruction part, we introduce an inpainting concentrated reconstruction loss that only computes the L1 distances within the occluded region. That is,
\begin{equation}
    \begin{split}
    \mathcal{L}^i_{rec} = \lVert y_i \odot (1 - M) -
    G^i(x, z) \odot (1 - M) \rVert_1.
    \end{split}
\end{equation}
With this scheme, we make sure that no errors outside the occluded region will be backpropagated to the generator, and therefore the unnecessary autoencoding is not learned. Figure \ref{fig: inpainting_concentration}(c) shows an example output of the network trained with the proposed scheme. The network produces sharp and realistic results inside the occluded region and produces inferior results for the context region that contributes little to the inpainting. Figure \ref{fig: inpainting_concentration}(d) is obtained using Equation (\ref{eq: replacing}). The inpainted region coherently fits with the context and the image overall looks realistic.

\subsection{Network Architecture} \label{sec: network_architecture}

For the generator and discriminator, we follow the architecture choices in \cite{radford2015unsupervised} for stable deep convolutional GANs. Both the generator and discriminator take an input image of size $128\times128$. For the generator, its encoder has $7$ convolution layers with each layer followed by a batch normalization layer \cite{ioffe2015batch} and a LeakyReLU \cite{he2015delving} layer.
The decoder has a symmetric structure with the encoder, except that it uses transposed convolution and ReLU \cite{krizhevsky2012imagenet}. All the convolutional layers have a $4\times4$ kernel size with a stride of 2. The output layer of the generator is a $\tanh$ function.
We also adapt the design suggestion from \cite{isola2016image} by adding skip connections between encoder and decoder. Skip connections shuttle low level features directly to decoder without passing through the ``bottleneck layers''. Such a circumvention is critical to some tasks such as semantic segmentation \cite{ronneberger2015u} (which is also included in the collaborative training) and we also find this helpful in improving the coherence between the context and the inpainted region. The discriminator has a similar structure to the encoder, except that it only has $5$ convolutional layers.

\section{Experimental Results}


\subsection{Datasets} \label{sec: datasets}

The dataset used in our experiment is the CelebA \cite{liu2015faceattributes} dataset. It has $202,599$ face images and we use the official split for training, validating and testing. Unlike the state-of-the-art works \cite{yeh2016semantic,li2017generative}, we do not align the faces according to the eyes. In fact, we find such alignment makes the inpainting much easier for the models as they do not need to semantically learn too much about the locations of the eyes and other face parts. Hence, to avoid overfitting and achieve better generalization, when cropping the faces, we only guarantee that the eyes, nose and mouth are included and \textit{no alignment is performed}. We also augment the dataset by random shift, scaling, rotation and flipping to further ensure the diversity of the faces during training.

Along with each face image, the CelebA dataset readily provides the locations of 5 face landmarks (the two eyes, nose and the two corners of mouth).
We create heatmaps from the landmarks according to the method denoted in \cite{pfister2015flowing,payer2016regressing}. The generated heatmaps will be used during collaborative training. For a fair comparison, we obtain segmentation masks for each of the faces using the parsing network provided by \cite{li2017generative}. The network is trained based on the Helen \cite{le2012interactive} dataset and achieves a close to state-of-the-art performance. Hence, it is considered sufficient for this study, which is supported by our experimental results. Also, using computer-generated annotations alleviates the burden of laborious manual annotation effort.


All images in the experiments are resized to $128\times128$. We use this size choice for a fair comparison with other approaches. It is straighforward for the proposed method to use a large image size. Following the mask generation strategies in previous works, we apply three different masks to the resized images: 1) random block mask with a $64\times64$ block \cite{li2017generative}; 2) random pattern mask \cite{pathak2016context} with roughly 25\% of the pixels missing; 3) random noise mask with  \textit{80\% of the pixels missing} \cite{yeh2016semantic}. For 1) and 3) the masked region is filled with random noise. For 2), the masked region is filled with zeros as the mask itself is already noisy.

\subsection{Models}

To demonstrate the effectiveness of the proposed method, the performances under different model settings are investigated. We denote $M_\Omega$ as the model trained using $L^{\Omega}$ and $M_{\Omega}^*$ as the model trained in addition with the inpainting concentrated scheme. Model settings are changed by varying $\Omega$ and switching between $M_\Omega$ and $M_{\Omega}^*$. All the investigated models are trained for 20 epochs. For the optimization, we use the Adam \cite{kingma2014adam} optimizer with a learning rate of $0.01$. For the weights of different losses, we emperically find the following settings work well:
\begin{itemize}
    \item Models with one task: $\lambda_{adv}^{i} = 1.0$, $\lambda_{adv}^{s} = 1.0$, $\lambda_{adv}^{d} = 1.0$, $\lambda_{res}^{i} = 100$, $\lambda_{res}^{s} = 1000$, $\lambda_{res}^{d} = 1000$;
    \item Models with two tasks: $\lambda_{adv}^{i} = 0.8$, $\lambda_{adv}^{s} = 0.2$, $\lambda_{adv}^{d} = 0.2$, $\lambda_{res}^{i} = 100$, $\lambda_{res}^{s} = 200$, $\lambda_{res}^{d} = 200$;
    \item Models with three tasks: $\lambda_{adv}^{i} = 0.8$, $\lambda_{adv}^{s} = 0.1$, $\lambda_{adv}^{d} = 0.1$, $\lambda_{res}^{i} = 100$, $\lambda_{res}^{s} = 200$, $\lambda_{res}^{d} = 200$.
\end{itemize}
For models with one or two tasks, the parameters will be used only when they are applicable. For example, model $\mathrm{M_{i}}$ will only have $\lambda_{adv}^{i} = 1.0$ and $\lambda_{res}^{i} = 100$. Other parameters are not used as they are for $\mathrm{M_{s}}$ and $\mathrm{M_{d}}$. Note for this study we are not interested in the best parameter settings for each model. Hence, the parameters are chosen when they work reasonably well. For the $\lambda_{adv}^{i}$, we generally find $0.8$ works better in a multi-task scenario and other adversarial loss paramters are chosen such that they sum up to 1. For the reconstruction loss parameters, we find minor performance differences when they are in a reasonable range. In general, setting $\lambda_{res}^{i}$ around $100$ and $\lambda_{res}^{s}$ and $\lambda_{res}^{d}$ around $200$ gives good performance when several tasks are presented. Setting values close to these numbers give similar performances and the performances are degraded when the values are too large or too small.


\subsection{Face Completion}

\begin{figure}[t]
    \centering
    \begin{subfigure}[b]{0.136\textwidth}
        \includegraphics[width=1.05\linewidth]{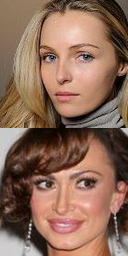}
        \vspace*{-14pt}
        \caption{original}
    \end{subfigure}
    \begin{subfigure}[b]{0.136\textwidth}
        \includegraphics[width=1.05\linewidth]{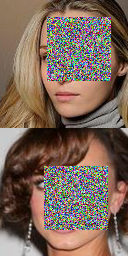}
        \vspace*{-14pt}
        \caption{masked}
    \end{subfigure}
    \begin{subfigure}[b]{0.136\textwidth}
        \includegraphics[width=1.05\linewidth]{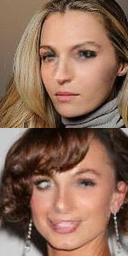}
        \vspace*{-14pt}
        \caption{$\mathrm{M_{i}}$}
    \end{subfigure}
    \begin{subfigure}[b]{0.136\textwidth}
        \includegraphics[width=1.05\linewidth]{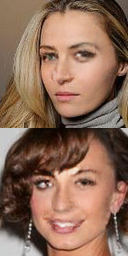}
        \vspace*{-14pt}
        \caption{$\mathrm{M_{i,d}}$}
    \end{subfigure}
    \begin{subfigure}[b]{0.136\textwidth}
        \includegraphics[width=1.05\linewidth]{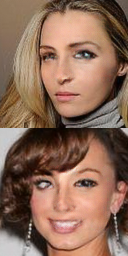}
        \vspace*{-14pt}
        \caption{$\mathrm{M_{i,s}}$}
    \end{subfigure}
    \begin{subfigure}[b]{0.136\textwidth}
        \includegraphics[width=1.05\linewidth]{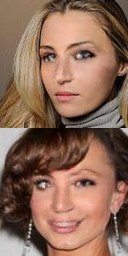}
        \vspace*{-14pt}
        \caption{$\mathrm{M_{i,s,d}}$}
    \end{subfigure}
    \begin{subfigure}[b]{0.136\textwidth}
        \includegraphics[width=1.05\linewidth]{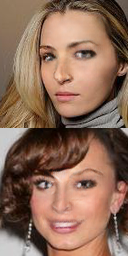}
        \vspace*{-14pt}
        \caption{$\mathrm{M_{i,s,d}^*}$}
    \end{subfigure}
    \caption{Qualitative face completion comparison of our models with different settings and varying numbers of tasks.}
    \label{fig: ours_comparison}
\end{figure}

\noindent \textbf{Qualitative Comparison} \hspace{5pt} During the experiments, we find in general the models trained with more collaborative tasks produce more realistic results. Figure \ref{fig: ours_comparison} shows some example outputs of our models. The first row demonstrates that, for an easy inpainting task where context coherence is not critical, the models can all produce relatively good looking faces and the collaboratively trained models from (d) to (g) tend to emphasis more on landmarks (eyes in this case) to give even better inpainting outputs. In the second row, we show a challenging case that only part of the woman's left eye is present and the entire right eye is missing. This inpainting task requires the model to both complete the left eye and reconstruct the entire right eye such that the two eyes together looking realistic. Without collaborative inpainting, the $\mathrm{M_{i}}$ model does not consider the coherence between the two eyes and inpaints the two eyes independently. The other models try to more or less balance the two eyes so that they have the same shape, size and color. The two models trained with three tasks in the last two columns give relatively better inpainting of the eyes. Overall, the $\mathrm{M_{i,s,d}^*}$ model trained using the inpainting concentrated scheme gives the most realistic output.

\begin{figure}[t]
    \centering
    \includegraphics[width=\linewidth]{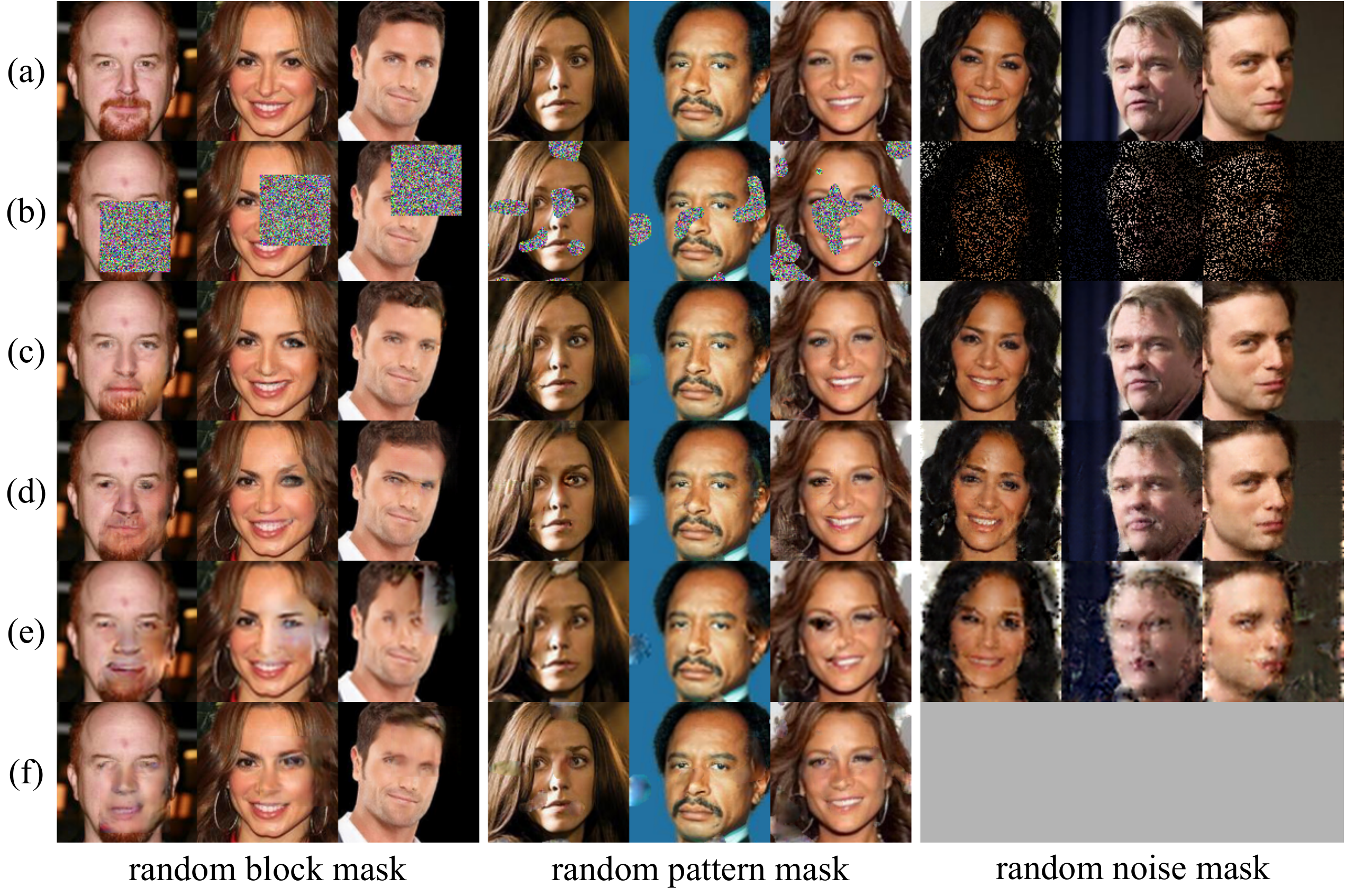}
    \caption{Qualitative face completion comparison between the proposed method and the state-of-the-art methods. Different masks are applied. From (a) to (f): (a) original image, (b) masked image, and results of (c) the proposed $\mathrm{M_{i,s,d}^*}$ model, (d) CE \cite{pathak2016context}, (e) SII \cite{yeh2016semantic} and (f) GFC \cite{li2017generative}.}
    \label{fig: others_comparison}
\end{figure}

We then compare the proposed method with other state-of-the-art models: CE \cite{pathak2016context}, SII \cite{yeh2016semantic} and GFC \cite{li2017generative}. Since we use a slightly different face cropping and data augmentation strategy, we retrain those models for a fair comparison. All retrainings are based on their officially released code with the training parameters unchanged. For SII \cite{yeh2016semantic}, all the experiments are performed using $64\times64$ images as DCGAN \cite{radford2015unsupervised} only works well on images with lower resolution. For all the models, poisson blending is performed. The comparison results are given in Figure \ref{fig: others_comparison}. To demonstrate the generalizability of the models on various shaped masks, the random pattern masked images in columns 4-6 are inpainted using the models trained with random block masks. For the random noise mask case, since GFC \cite{li2017generative} requires square masks for the local loss, it cannot be used to complete random noise masked images. Hence, the inpainting results for GFC \cite{li2017generative} in columns 7-9 are omitted. Also, for CE \cite{pathak2016context} and our method, we train a new model for the random noise mask case due to the uniqueness of the mask.

From Figure \ref{fig: others_comparison}, we observe that our model consistently gives better inpainting results than the state-of-the-art methods. In general, SII \cite{yeh2016semantic} gives the worst results in all the cases. Due to the unaligned faces and the data augmentation we performed during training, the face scene becomes more complex. Thus, the DCGAN model used in SII \cite{yeh2016semantic} can not learn the face data distribution very well, which as a result yieds inferior inpainting performance during the inference step. The CE \cite{pathak2016context} and GFC \cite{li2017generative} models in general produce reasonably good results, especially when the faces are aligned. However, as they either train the model without using additional structural constrains or has an indirect measurement of structural inconsistency, their synthesized images are less structurally realistic than ours.

\begin{figure}[t]
    \centering
    \begin{subfigure}[b]{0.105\textwidth}
        \includegraphics[width=1.01\linewidth]{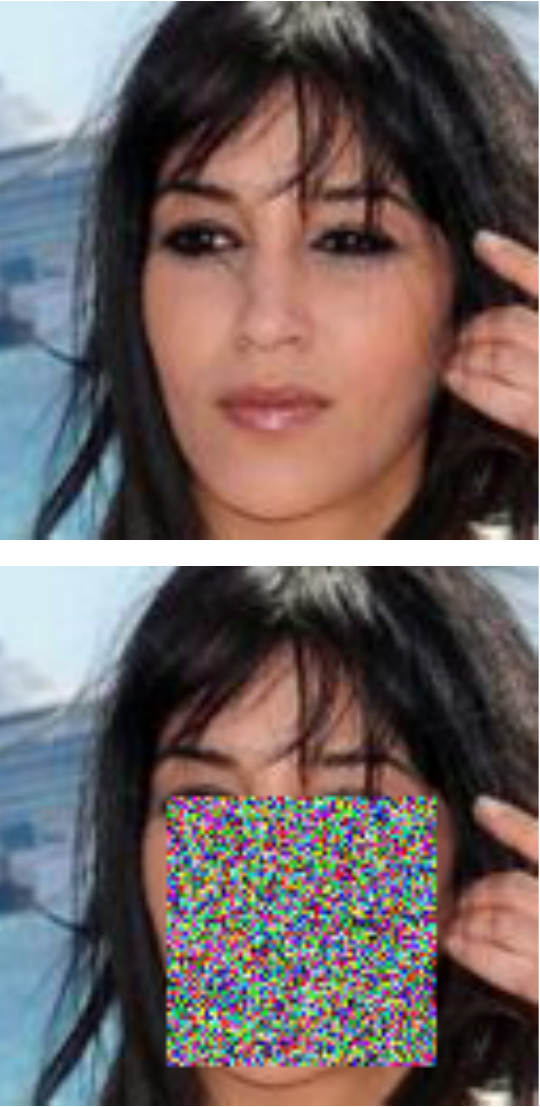}
        \vspace*{-14pt}
        \caption{}
    \end{subfigure}
    \begin{subfigure}[b]{0.428\textwidth}
        \includegraphics[width=1.01\linewidth]{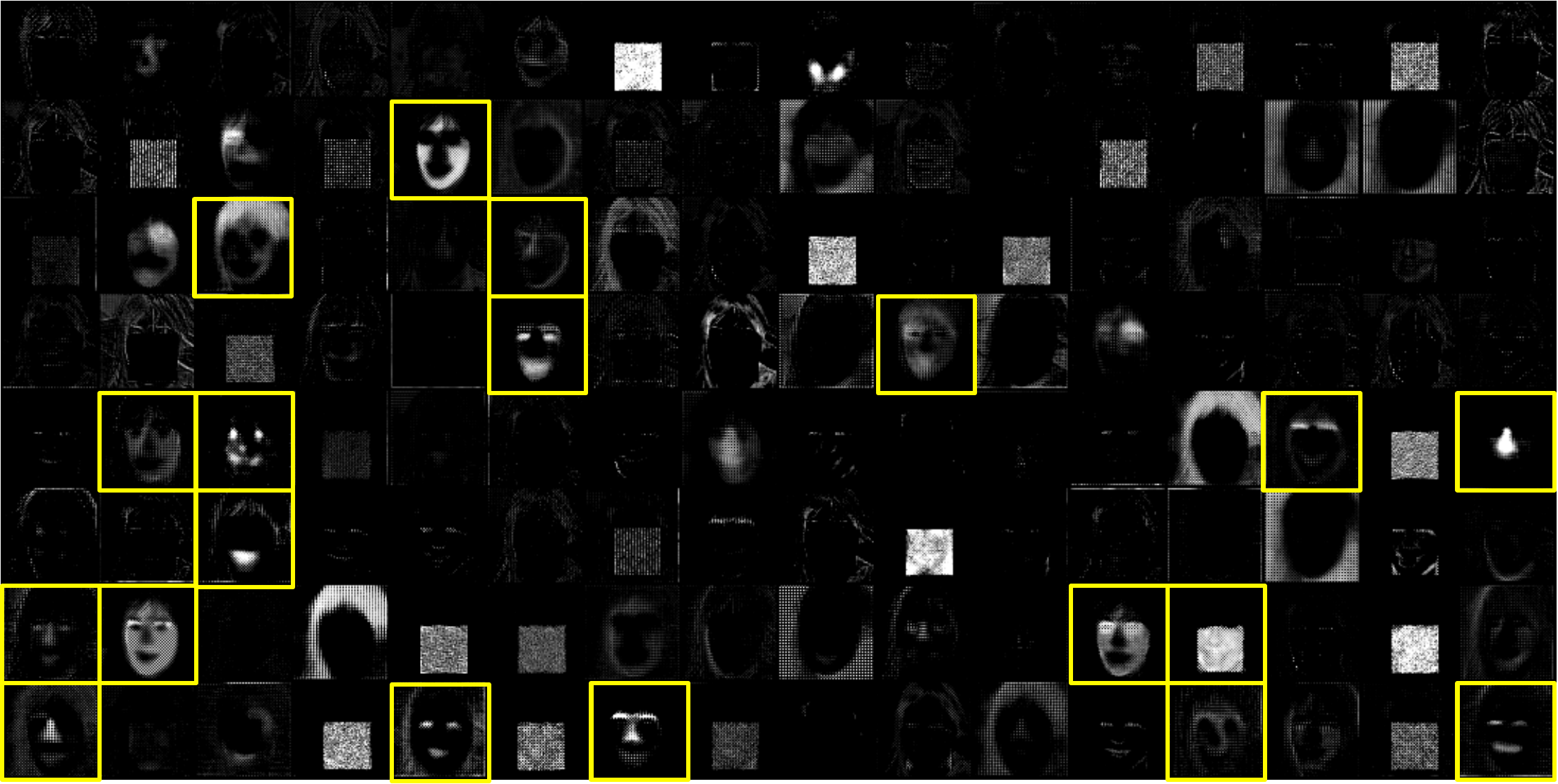}
        \vspace*{-14pt}
        \caption{}
    \end{subfigure}
    \begin{subfigure}[b]{0.428\textwidth}
        \includegraphics[width=1.01\linewidth]{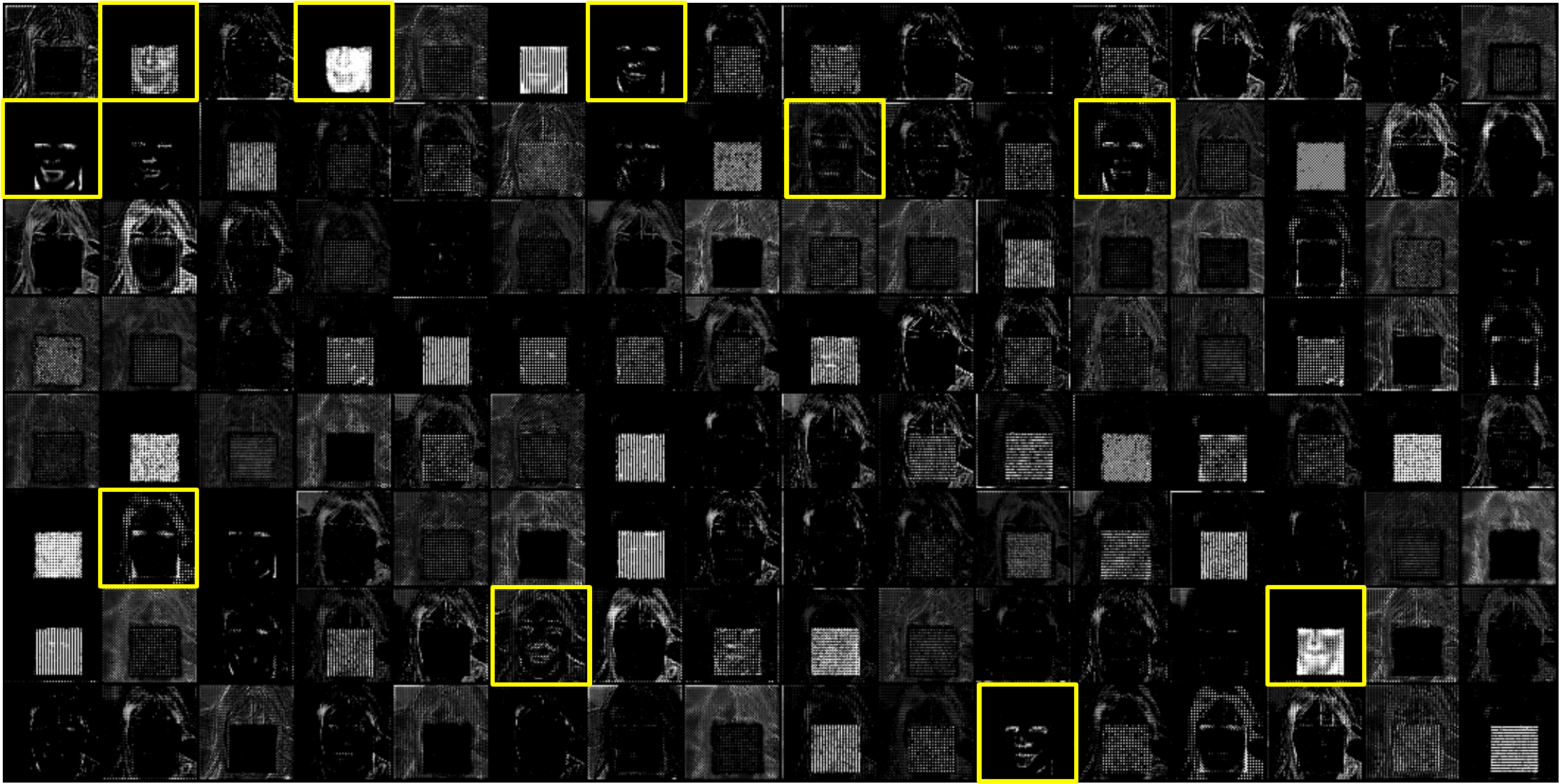}
        \vspace*{-14pt}
        \caption{}
    \end{subfigure}
    \caption{Feature maps from the last common layer in the generator G. Face related features are marked with yelow squares. Features from early layers through the skip connection are omitted. (a) The original and masked images, (b) Feature maps of the $\mathrm{M_{i,s,d}}$ model, and (c) Feature maps of the $\mathrm{M_{i}}$ model.}
    \label{fig: features}
\end{figure}

Figure \ref{fig: features} shows the feature maps extracted from the last common layer in the generator G of the $\mathrm{M_{i,s,d}}$ and $\mathrm{M_{i}}$ models. The $\mathrm{M_{i,s,d}}$ model is more predictive in the masked region
and outputs better face related features. It is interesting to notice that the $\mathrm{M_{i,s,d}}$ model tends to treat face parts independently. Some maps contain features only about nose, mouse or eyes. While the $\mathrm{M_{i}}$ model mostly outputs features about the whole face. This indicates that the $\mathrm{M_{i,s,d}}$ model is more discriminative about the face than the $\mathrm{M_{i}}$ model and it learns to distinguish (and generate) each of the face part due to the training with other tasks.

\vspace{2mm}

\noindent \textbf{Quantitative Comparison} \hspace{5pt} In addition to the visual comparison, we also quantitatively compare our models with the state-of-the-art models to statistically understand their inpainting performances. All the evaluated models are trained with random block masks. We use two classic metrics, PSNR and SSIM, to evaluate the similarity between the inpainted image and the ground truth. PSNR gives the similarity score at pixel-level while SSIM evaluates the similarity at perceptual level. The ground truth image used in this evaluation is the original image before the occlusion. However, since given an occluded image, there could be multiple inpainting results that are perceptually correct, we realize that neither PSNR or SSIM offers a perfect evaluation. But due to the symmetric nature and the relatively simple structure of human faces, as long as the occlusion does not hide significant portion of the face, all the possible inpainting results should be similar which gives us an opportunity to roughly evaluate the models' performance. Table \ref{tab: quantitative_face} shows the evaluation results at 6 different mask locations. We adapt the mask location choices from \cite{li2017generative} by masking out left eye (O1), right eye (O2), upper face (O3), left face (O4), right face (O5), and lower face (O6), respectively. We intentionally select relatively smaller mask sizes (on average $40\times48$) to limit the variation of all possible inpaintings. We observe from Table \ref{tab: quantitative_face} that our base model $\mathrm{M_{i}}$ has already obtained comparable or slightly better performance than the state-of-the-art models which demonstrates the effectiveness of introducing skip connections into collaGAN. In general, models trained with more tasks give better numbers in both PSNR and SSIM. This shows that the knowledge among tasks is indeed collaborative during training. We also find that $\mathrm{M_{i,s,d}^*}$ performs slightly better than $\mathrm{M_{i,s,d}}$ which means the inpainting concentrated scheme is really helpful in getting better inpainting results. Overall, we spot a significant performance jump with the proposed method when compared with the state-of-the-art models.

To further understand the collaborative nature of the proposed method, we also investigate if the inpainting can help other tasks as well. Specifically, we want to know that, through the learning of inpainting, if the model can predict better of the semantical labels and landmarks in the occluded region. We use the Dice coefficient and localization error to evaluate the performance of semantic segmentation and landmark detection, respectively. The Dice coefficient measures the similarity of two segmentation masks and the localization error computes the Euclidean distances between two landmarks. Table \ref{tab: segmentation} gives the Dice coefficient of different models trained with the segmentation task. The performances of the facial semantic labels are shown.
It is clear that \textit{the collaboratively trained models perform better than single task models and the models trained with three tasks better than with two tasks}. The $\mathrm{M_{i,s,d}}$ and $\mathrm{M_{i,s,d}^*}$ models have very close performance. This is reasonable as the inpainting concentrated scheme is designed solely for inpainting and does not introduce additional information to other tasks during training. Note that, for the segmentation task, the performance difference of the models is not significant. We speculate that this is because the ``ground truth" is computer-generated from the parsing network which sometimes may not give perfect prediction. For the landmark detection task with accurately labeled landmark locations provided by the CelebA dataset, the performance boost of introducing other tasks is more significant as shown in Table \ref{tab: detection}. For example, the average landmark error is reduced from 2.38 (the $\mathrm{M_{d}}$ model) to 1.72 (the $\mathrm{M_{i,s,d}^*}$ model), a 27.7\% decrease. 

\begin{table}[t]
    \centering
    \caption{Quantitative face completion comparison of different models evaluated at 6 different mask locations: left eye (O1), right eye (O2), upper face (O3), left face (O4), right face (O5), and lower face (O6). The numbers in each cell are SSIM (\%)/PSNR (dB), the higher the better.}
    \label{tab: quantitative_face}
    \resizebox{\textwidth}{!}{
        \begin{tabular}{@{}rcccccccccccc@{}}
            \toprule
                         &  & \multicolumn{1}{c}{\scriptsize \textbf{O1}} & \multicolumn{1}{c}{\textbf{}} & \multicolumn{1}{c}{\scriptsize \textbf{O2}} & \multicolumn{1}{c}{\textbf{}} & \multicolumn{1}{c}{\scriptsize \textbf{O3}} & \multicolumn{1}{c}{\textbf{}} & \multicolumn{1}{c}{\scriptsize \textbf{O4}} & \multicolumn{1}{c}{\textbf{}} & \multicolumn{1}{c}{\scriptsize \textbf{O5}} & \multicolumn{1}{c}{\textbf{}} & \multicolumn{1}{c}{\scriptsize \textbf{O6}} \\ \midrule
            \scriptsize{\textbf{CE} \cite{pathak2016context}}  &  & 90.5/26.74                      &                               & 90.6/27.01                      &                               & 93.8/27.90                      &                               & 95.8/30.37                      &                               & 96.0/30.65                      &                               & 90.0/27.11                      \\
            \scriptsize{\textbf{SII} \cite{yeh2016semantic}} &  & 87.5/23.93                      &                               & 87.7/24.12                      &                               & 93.0/26.61                      &                               & 95.6/28.94                      &                               & 95.9/29.38                      &                               & 87.8/24.84                      \\
            \scriptsize{\textbf{GFC} \cite{li2017generative}} &  & 90.6/27.10                      &                               & 90.9/27.34                      &                               & 94.0/28.68                      &                               & 96.3/31.18                      &                               & 96.3/31.11                      &                               & 90.0/27.13                      \\
            \scriptsize{$\mathbf{M_{i}}$}       &  & 90.8/27.23                      &                               & 91.1/27.42                      &                               & 94.0/28.53                      &                               & 96.0/30.94                      &                               & 96.1/31.06                      &                               & 90.1/27.15                      \\
            \scriptsize{$\mathbf{M_{i,d}}$}     &  & 91.7/27.57                      &                               & 91.6/27.81                      &                               & 94.5/28.76                      &                               & 96.4/31.33                      &                               & 96.5/31.36                      &                               & 90.7/27.37                      \\
            \scriptsize{$\mathbf{M_{i,s}}$}     &  & 91.4/27.59                      &                               & 91.5/27.65                      &                               & 94.3/28.66                      &                               & 96.4/31.22                      &                               & 96.4/31.28                      &                               & 90.7/27.55                      \\
            \scriptsize{$\mathbf{M_{i,s,d}}$}   &  & 91.7/27.66                      &                               & 91.8/27.87                      &                               & 94.5/28.77                      &                               & 96.5/31.31                      &                               & 96.6/31.39                      &                               & 90.8/27.57                      \\
            \scriptsize{$\mathbf{M^*_{i,s,d}}$}  &  & \textbf{92.4/27.76}             & \textbf{}                     & \textbf{92.6/27.96}             & \textbf{}                     & \textbf{95.2/28.79}             & \textbf{}                     & \textbf{97.2/31.44}             & \textbf{}                     & \textbf{97.2/31.50}             & \textbf{}                     & \textbf{91.7/27.81}             \\ \bottomrule
        \end{tabular}
    }
\end{table}

\begin{table}[t]
    \centering
    \begin{minipage}[t]{.47\linewidth}
        \centering
        \caption{Semantic segmentation performance of different models. The numbers are given as Dice coeffient (\%). Higher numbers are better.}
        \label{tab: segmentation}
        \begin{tabular}{@{}rccccccc@{}}
        \toprule
        & & \scriptsize{$\mathbf{M_{s}}$} & & \scriptsize{$\mathbf{M_{i,s}}$} & & \scriptsize{$\mathbf{M_{i,s,d}}$}   & \scriptsize{$\mathbf{M_{i,s,d}^*}$}    \\ \midrule
        \scriptsize{\textbf{Face}}          & & 93.7            & & 94.1       & & \textbf{94.2} & 94.1          \\
        \scriptsize{\textbf{Left eyebrow}}  & & 74.2            & & 74.6       & & 75.0          & \textbf{75.2} \\
        \scriptsize{\textbf{Right eyebrow}} & & 72.3            & & 72.8       & & \textbf{73.8} & 73.5          \\
        \scriptsize{\textbf{Left eye}}      & & 70.7            & & 71.9       & & 72.2          & \textbf{72.7} \\
        \scriptsize{\textbf{Right eye}}     & & 70.0            & & 70.3       & & \textbf{71.2} & \textbf{71.2} \\
        \scriptsize{\textbf{Nose}}          & & 90.5            & & 90.9       & & 90.9          & \textbf{91.0} \\
        \scriptsize{\textbf{Upper lip}}     & & 68.1            & & 67.2       & & \textbf{68.3} & 67.9          \\
        \scriptsize{\textbf{Teeth}}         & & 64.2            & & 66.5       & & \textbf{66.8} & 66.7          \\
        \scriptsize{\textbf{Lower lip}}     & & 81.4            & & 82.1       & & \textbf{82.8} & 82.4          \\ \bottomrule
        \scriptsize{\textbf{Average}}       & & 76.1            & & 76.7       & & \textbf{77.2} & \textbf{77.2}          \\ \bottomrule
        \end{tabular}
    \end{minipage}%
    \hspace{1em}
    \begin{minipage}[t]{.47\linewidth}
        \centering
        \caption{Landmark detection performance of different models. The numbers are given as localization errors in pixels. Lower numbers are better.}
        \label{tab: detection}
        \begin{tabular}{@{}rccccccc@{}}
        \toprule
                             & & \scriptsize{$\mathbf{M_{d}}$} & & \scriptsize{$\mathbf{M_{i,d}}$} & & \scriptsize{$\mathbf{M_{i,s,d}}$}   & \scriptsize{$\mathbf{M_{i,s,d}^*}$}   \\ \midrule
        \scriptsize{\textbf{Left eye}}    & & 2.12             & & 1.73        & & 1.61          & \textbf{1.60} \\
        \scriptsize{\textbf{Right eye}}   & & 2.29             & & 1.71        & & \textbf{1.59} & 1.62          \\
        \scriptsize{\textbf{Nose}}        & & 2.53             & & 2.15        & & \textbf{1.92} & 1.93          \\
        \scriptsize{\textbf{Left mouth}}  & & 2.39             & & 1.80        & & 1.74          & \textbf{1.73} \\
        \scriptsize{\textbf{Right mouth}} & & 2.57             & & 1.86        & & 1.77          & \textbf{1.72} \\ \bottomrule
        \scriptsize{\textbf{Average}}     & & 2.38             & & 1.85        & & 1.73          & \textbf{1.72} \\ \bottomrule
        \end{tabular}
    \end{minipage} 
\end{table}

\section{Conclusion}
    
We present a novel collaborative GAN framework for face completion. The experimental results suggest that training multiple related tasks together within the proposed framework is beneficial. By infusing more knowledge, the generative model learns better about the inpainting through the knowledge sharing of the segmentation and detection tasks, whose performances are boosted vice versa. We have also found that optimizing directly toward inpainting other than autoencoding produces better inpainting results in an image-to-image network. Finally, we have demonstrated that the proposed method can give superior inpainting performance than the state-of-the-art methods.
\vspace{1em}

\noindent \textbf{Disclaimer:} This feature is based on research, and is not commercially available. Due to regulatory reasons its future availability cannot be guaranteed.

\bibliographystyle{splncs04}
\bibliography{references}

\end{document}